\documentclass[copyright,creativecommons]{eptcs}
\providecommand{\event}{ICLP 2025} 

\usepackage{iftex}

\ifpdf
  \usepackage{underscore}         
  \usepackage[T1]{fontenc}        
\else
  \usepackage{breakurl}           
\fi

\usepackage{graphicx}
\usepackage{subcaption}

\usepackage{algorithm}
\usepackage[indLines=true,noEnd=true,italicComments=true]{algpseudocodex}
\usepackage{fancyvrb} 
\usepackage{fvextra}

\newcommand{\smodels}{\models_{sm}}
\newcommand{\naf}{{\mathtt{not}}\,}
\newcommand{\False}{{\mathtt{False}}}
\newcommand{\True}{{\mathtt{True}}}
\newcommand{\set}[2]{\{ {#1} \,|\,  {#2}\}}
\newcommand{\codeif}{\mathbin{:\!-\,} }

\title{XAI-LAW: A Logic Programming Tool for Modeling, Explaining, and Learning Legal Decisions~\thanks{Research partially
supported by Interdepartment Project on AI (Strategic Plan of UniUD-22-25)}}
\author{Agostino Dovier \qquad Talissa Dreossi \qquad Andrea Formisano \qquad Benedetta Strizzolo
\institute{DMIF -- University of Udine,  Udine, Italy}
\institute{Gruppo Nazionale per il Calcolo Scientifico -- INdAM, Roma, Italy}
\email{agostino.dovier@uniud.it, talissa.dreossi@uniud.it} \email{andrea.formisano@uniud.it, strizzolo.benedetta@spes.uniud.it}
}

\def\titlerunning{XAI-LAW: A Logic Programming Tool for Modeling, Explaining, and Learning Legal Decisions}
\def\authorrunning{A. Dovier, T. Dreossi, A. Formisano \& B. Strizzolo}

\begin{document}
\maketitle

  \begin{abstract}
We propose an approach to model articles of the Italian Criminal Code (ICC), using Answer Set Programming (ASP), and to semi-automatically learn legal rules from examples based on prior judicial decisions.
The developed tool is intended to support legal experts during the criminal trial phase by providing reasoning and possible legal outcomes. 
The methodology involves analyzing and encoding articles of the ICC in ASP, including ``crimes against the person'' and property offenses.
The resulting model is validated on a set of previous verdicts and refined as necessary.
During the encoding process, contradictions may arise; these are properly handled by the system, which also generates possible decisions for new cases and provides explanations through a tool that leverages the ``supportedness'' of stable models. The automatic explainability offered by the tool can also be used to clarify the logic behind judicial decisions, making the decision-making process more interpretable.
Furthermore, the tool integrates an inductive logic programming system for ASP, which is employed to generalize legal rules from case examples.
\end{abstract}

\section{Introduction}

The desire of delegating legal decisions to an automated formal system can be traced back (at least) to Leibniz's dream. In spite of G\"odel incompleteness results, holding in a more general context, several efforts have been posed in this direction since the birth of artificial intelligence, earlier expert systems, and logic programming.
From the 1950s, Allen and Saxon~\cite{Allen57,DBLP:conf/icail/AllenS87} paved the way for the application of logic programming in the legal field, by interpreting legal documents, and by outlining what is necessary for encoding legal rules in logic. Two of the main challenges in this area concern the inherent ambiguity of \emph{natural language} (NL) and the applicability of legal rules.

The first notable attempt to encode laws in logic was undertaken by Kowalski et al. \cite{10.1145/5689.5920,kowalski1992legislation,kowalski2023logical}, 
who formalised part of the British Nationality Act. They introduced a specialised language, called Logical English, which consists of a \emph{controlled natural language} (CNL), i.e. a language that easily allows English sentences to be translated into logic programs such as Prolog, Datalog, or Answer Set Programming (ASP) \cite{kowalski2023logical}. 
In this field, progress has been made through the development of systems that automatically translate NL into CNL and then into ASP code \cite{borroto2024automaticcompositionaspprograms}. Similarly, Large Language Models (LLMs) can be employed to directly translate NL into ASP  \cite{ricca1,coppolillo2024}. 
A further step was to combine LLM-based semantic parsers with a learning module, such as in the aforementioned paper by Kareem et al.~\cite{ricca1}. In that study, ILASP, which is also used in this paper, was employed to learn the ASP programs generated by LLMs. While LLMs are valuable tools for processing text and converting it into corresponding logical encodings, their primary limitation lies in the granularity of the generated encodings. In particular, a common challenge with legal rules is the occurrence of distinct terms that share the same meaning. To address this issue, it is necessary to either distinguish between or unify these terms based on their meaning, as well as establish possible co-references within the text.

At the beginning Prolog was the language employed in legal reasoning; however due to its limited capabilities on commonsense/non-monotonic reasoning and in particular in handling exceptions, belief  revision, default negation, ASP offers a more appropriate alternative~\cite{baral1994logic}. 
As an example, Athakravi et al.~\cite{Athakravi2015} proposed a logic-based approach to generate a declarative model of judgments from past legal cases, capturing legal reasoning through rules and exceptions. In contrast, s(CASP), 
used by Sartor et al.~\cite{DBLP:conf/iclp/SartorDBCPK22}, follows a top-down goal-oriented strategy. It includes an explanation tool that traces the paths leading to the solution; however,
it does not provide a comprehensive model of the ASP encoding. Top-down approaches, however, are generally less suitable for legal reasoning due to scalability issues and may yield to conclusions even when applied to inconsistent programs, i.e. those that do not admit any stable model. 
Similarly, the tool developed by Arias et al.~\cite{arias2024automated} uses s(CASP), and hence it must operate under the assumption that the system is consistent; as a matter of fact, it may return an answer even when the program does not admit any stable model. 

One key advantage of ASP is its explainability in contrast to black-box algorithms, which present four different issues: opacity, strangeness, unpredictability, and justification~\cite{brozek2024black}. Notably, even individuals involved in the penal system, such as police officers, have expressed scepticism towards opaque tools~\cite{simmler2023smart}. An example of a widely discussed black-box system in the legal domain is COMPAS (Correctional Offender Management Profiling for Alternative Sanctions) \cite{kirchner2018compas}, a proprietary risk assessment tool used in the United States to predict the likelihood of criminal reoffending. It has been heavily criticised for its lack of transparency and potential biases, particularly regarding race and socio\-economic status.

In this context, Inductive Learning of Answer Set Programs (ILASP), an \emph{Inductive Logic Programming} (ILP) framework \cite{law2020ilasp}, offers a novel approach for learning legal reasoning, while ensuring interpretability. The reason why we chose this approach is that it employs ASP, allowing the development of a non-monotonic reasoning system. In this study, we exploit ASP, for encoding immutable legal rules, and a {Conflict-Driven Clause Learning} (CDCL) bottom-up ASP solver, to derive stable models. 
In previous work \cite{TALUPolonia24} we focused on 
the law encodings in particular on how dealing the notion of vagueness; in~\cite{cilc2024} the basis of the project later developed for this paper were presented. In this work, we finalized that studies by augmenting the set of encoded articles from the Italian penal code, applying ILASP (version~4) to derive rules explaining judges' reasoning patterns. 
Moreover, since ASP by itself produces stable models but does not show the rules that were triggered to get those results, as explained by \cite{Robaldo2023}, in this paper we propose the use of xASP \cite{DBLP:journals/logcom/AlvianoTSB24}. XASP is a tool that allows us to interpret and understand the reasoning behind the stable models generated by the ASP solvers by constructing a \emph{Directed Acyclic Graph} (DAG). The integration of ILP framework and xASP has already been proved to work in a similar study applied to a different field \cite{dreossi2024towards}. This approach enables the system to compute scenarios of plausible decisions, each accompanied by an explanation, which can then be reviewed by a judge to inform her/his final decision. Alternatively, it can be used to generate a logical explanation for decisions autonomously made by a judge.
Although, in principle, the system could be used for autonomous decision-making, the authors would suggest its use as a decision support system.

\smallskip

The paper is organized as follows. In Section~\ref{sect:prelim} we briefly recall the basic notions on ASP and ILP.
Section~\ref{sect:ASPLAW} describes the main aspects of our ASP encoding of the Italian Criminal Code. Such encoding forms the basis for the tasks of prediction, explanation, and dynamic model refinement described in Sections~\ref{sect:predict} and~\ref{sect:medelRefine}.  In Section~\ref{Conclusions} we briefly present the overall system and we draw some conclusions and lines for future work.

\section{Preliminaries}\label{sect:prelim}

We assume the reader is aware of the main logic programming concepts. We repeat here some of the
definitions for the sake of readability.

\subsection{ASP and Stable Model Semantics}
Let us consider a denumerable set of variables $\cal X$, a set of constant symbols
$\cal C$, and a set of predicate symbols $\cal P$.
An atomic formula (atom) is a
formula of the form 
$p(t_1,\dots,t_n)$, where $p \in {\cal P}$, and $t_1,\dots,t_n$ are constant symbols or variables. A literal is either an atomic formula or $\naf~A$, where $A$ is an atomic formula.
A (general) \emph{rule}  is a formula of the form 
\begin{equation}\label{eq:rule}
 H \leftarrow   B_1,~\dots,~B_m,~\naf~C_1,~\dots,~\naf ~C_n
\end{equation}  
where $H, B_i, C_j$ are atomic formulas. 
If $n=0$, a rule is said \emph{definite}; if $m=n=0$, it is a \emph{fact}.

Variables are implicitly universally quantified and, logically, rules are implications.
A \emph{program} $P$ is a set of clauses. 
A literal, a rule, a program is \emph{ground} if it contains no variables.
Given a rule,  its \emph{grounding} is the set of all ground rules obtained replacing consistently 
all its variables with constant symbols.
The grounding of a program $P$ is the set of all ground rules obtained by grounding rules in~$P$.

The semantics of a program is based on the notion of \emph{answer set/stable model}~\cite{gelfond1988stable}.
Given a ground program $P$, a set of atoms $M$ is an answer set of $P$ if it is the unique minimum model
of the so-called ``reduct'' $P^M$ obtained from $P$ by first removing all rules with $\naf  A$ in their body where $A \in M$ and 
then removing  the negated literals from all remaining rules (let us remark that $P^M$ is made by definite rules). A program may have no answer sets, one answer set,
or multiple answer sets. If $P$ admits a stable model we say that it is consistent.
We define with $\smodels$ the logical consequence under stable models, namely we 
denote with $P \smodels A$ the fact that $P$ is consistent and the atom $A$ is true in all stable models of $P$,
and with $P \smodels \naf A$ the fact that $P$ is consistent and the atom $A$ is false in all stable models of $P$.

In addition to the rules of form~(\ref{eq:rule}), the ASP language has been extended over time with several constructs such as denials (rules without head $H$, that state that the body must be false in all stable models), 
cardinality constraints, and aggregates that will be used in the following section. Of particular interest are the choice rules ($\texttt{lb}\{A_1;\dots;A_k\}\texttt{ub} \leftarrow   B_1,~\dots,~B_m,~\naf~C_1,~\dots,~\naf ~C_n$.), which are rules that use disjunction in the head. The number of satisfied atoms in their head matches the specified cardinality bounds (\texttt{lb} and \texttt{ub}). For example, consider the following ASP program $P$ constituted by just a single choice rule: \verb|1{p;q}2|. The answer sets of $P$ are: \{\verb|p|\}, \{\verb|q|\} and \{\verb|p|, \verb|q|\}. 
We refer the reader to the literature on ASP for a detailed treatment~\cite{DovierFP18,ASPBook,Lifschitz19book}.

Although deciding whether a ground program $P$ admits answer sets or not is NP-complete, efficient solvers for ASP  are available.
These tools search for answer sets of a given program $P$ by proceeding in two phases. First, they compute the grounding of $P$, and then they
look for answer sets of the ground version using several clever techniques such as conflict-driven clause (rule) learning \cite{ASPBook}.

\subsection{Inductive Logic Programming}\label{sectionNOPL}

Inductive Logic Programming, briefly ILP~\cite{DBLP:journals/ngc/Muggleton91}, is a subfield of both machine learning and logic programming, that focuses on learning logical rules from examples and background knowledge. 
The goal is to find a hypothesis (i.e., a set of rules) that explains the given  
examples within the context of the background knowledge (again, a set of rules). 
\begin{quote}
\it From inductive machine learning, ILP inherits its goal: to develop tools and techniques to induce hypotheses from observations (examples) and to synthesize new knowledge from experience. By using computational logic as the representation mechanism for hypotheses and observations, ILP can overcome the two main limitations of classical machine learning~[\ldots]
\begin{enumerate}
\item the use of a limited knowledge representation formalism [\ldots], and
\item difficulties in using substantial background knowledge in the learning process. 
\\~\hspace*{\fill}{S.H.~Muggleton} and {L.~De~Raedt}~\cite{DBLP:journals/jlp/MuggletonR94}
\end{enumerate}
\end{quote}

Logic Programming and in general Computational Logic focuses on deductive reasoning. From facts and rules, their logical consequences are computed. 
The activity of finding hypotheses, instead,  has to do with induction rather than deduction:
\begin{quote}
\it Induction is, in fact, the inverse operation of deduction, and cannot be conceived to exist without the corresponding operation, so that the question of relative importance cannot arise.
Who thinks of asking whether addition of subtraction is the more important process in arithmetic? But at the same time much difference in difficulty may exist between a direct and inverse operation. 
\hfill W.S. Jevons \cite{JEVONS}
\end{quote}

\subsection{ILP Learning tasks}

A \emph{learning problem (learning task)} is defined as a triple  $T= \langle B, S, E\rangle$, where 
\begin{enumerate}
\item $B$ is an ASP program called \emph{background knowledge} 
\item $S$ is a set of ASP rules called \emph{hypothesis space},  and
\item $E$ is a set of \emph{examples}. 
\end{enumerate}

This general schema admits some variants, in particular in the form of rules admitted in $B$ and in $E$ and
in the set of examples (that can be positive, negative, restricted only to facts or involving more general rules).
The overall idea is that of finding a  (minimal)  subset $H \subseteq S$ such that
\begin{eqnarray}\label{explain}
B \cup H  \mbox{~``explains''~the examples in } E
\end{eqnarray}
The informal notion \emph{explains} should be made explicit, but it depends on the kind of
rules admitted for $B$, $S$,   on the kind of examples considered, and on our requirements for 
a \emph{good} explanation.

\smallskip

ASP was formally defined in the late Nineties \cite{DBLP:books/sp/99/MarekT99},
 in particular with the availability of the first ASP solver {\sc Smodels}~\cite{DBLP:conf/lpnmr/NiemelaS97}.
Before that, ILP focused on definite clause programs (basically, the same admitted by the 
Prolog language~\cite{DBLP:conf/ifip/Kowalski74}). Let us focus directly into the 
more complex case of (general) ASP programs where a \emph{learning problem}  $T= \langle B, S, E\rangle$ is 
composed by $B$ and $S$ (general) ASP programs, 
and $E$ is made by two parts: $E^+$ which is a set of ground facts that should be entailed, and $E^-$ which is a set of ground facts that should be not.
The lack of monotonicity of the language of general ASP programs leaves (at least) two formal
approaches for the informal notion of \emph{explanation}.
\begin{description}
\item[Brave Reasoning]
We look for a subset $H \subseteq S$ such that there is \emph{at least one} stable model $M$ of $B \cup H$ such that
$E^+ \subseteq M \mbox{ and } E^- \cap M = \emptyset$.
In practice, we look for a program where at least one of its stable models is coherent with both the positive and the negative requirements
coming from the known examples $E$, namely a stable model of $B \cup H \cup \hat E$ where
$\hat E = \set{ \codeif \naf A}{A \in E^+}
\cup
\set{\codeif A}{A \in E^-}$ is a set of denials.
\item[Cautious Reasoning] 
We look for a subset $H \subseteq S$ such that \emph{for all stable models} $M$ of $B \cup H$ it holds that
$E^+ \subseteq M$ and $E^- \cap M = \emptyset$.
Using the notion of logical consequence under stable models, this can be restated as $B \cup H \smodels E$.
This can be computed calling a complete ASP solver {\tt Solver} that returns a (different) stable model per call, 
and if there is not a (another) stable model, as sketched in Algorithm 1.
\end{description}

\begin{algorithm}[tb]\label{algo4}
\caption{Cautious computation of $H$, given $B$ and $E = \langle E^+, E^- \rangle$}\label{alg:cap}
{\small\begin{algorithmic}
\ForAll{$H \in \wp(S)$} \Comment{Start from $\emptyset$ and add elements} 
\State $C \gets \True$
\Repeat
\State $M \gets  \Call{\tt Solver}{B \cup H}$
\If {$(M \neq \False) \wedge ((E^+ \not\subseteq M) \vee (E^- \cap M \neq \emptyset))$}
\State $C \gets \False$ 
\EndIf
\Until {$M = \False \vee C = \False$}
\If {$C$} \State \Return $H$ 
\EndIf
\EndFor 
\State \Return {$\False$}  
\end{algorithmic}
}\end{algorithm}

\subsection{ILASP}\label{SUBILASP}
ILASP~\cite{law2020ilasp} is a learning framework capable of solving more complex learning tasks than those needed for the scope of this paper. In particular, examples may be context-dependent, and errors in the input data (noisy data) can be handled with a notion of penalty.

Each encoded example in ILASP consists of three sets of ground atoms: inclusions, exclusions, and context. Considering the context \textit{C}, every example has a broader background knowledge. In particular, $B \cup H  \cup C$ must $cover$ all examples. Given any framework $ILP_X$, learning task $T_X$ and example $e$, we say that $e$ is covered by a hypothesis if the hypothesis meets all conditions that $T_X$ imposes on $e$. Specifically, at least one answer set must exist for each example, containing all inclusions while omitting all exclusions.

To describe the hypothesis space, various approaches can be employed. The first one is \emph{direct rule declaration}, which explicitly specifies rules to define the hypothesis space:
each rule in the list must be preceded by its \emph{length}, representing the number of literals or components that make up the rule, followed by the symbol $\sim$, indicating the belonging of this rule to the hypothesis space.
The second approach is \emph{mode declarations}: ground atoms that contain placeholders such as \texttt{var(t)} or \texttt{const(t)}, that can be substituted with any variable or constant of type \texttt{t}, respectively.
To declare head predicates, the notation \texttt{\#modeh(pred(.))} is used, or alternatively, \texttt{\#modeha(pred(.))} if aggregates are permitted. For body predicates, \texttt{\#modeb(n, pred(.))} is employed, where \texttt{n} defines the maximum number of times \texttt{pred(.)} can appear in the body. Additionally, conditions can be specified using \texttt{\#modec(c)}, where \texttt{c} represents a logical condition, such as \texttt{var(t1)>var(t2)}. 
To reduce the hypothesis space size and the time spent to compute it, specialized constants can be introduced within the different kinds of mode declarations. The constant \texttt{positive} restricts the search by excluding negation as failure of the predicate from the mode declaration. 
Additional constraints, such as \texttt{symmetric} and \texttt{anti\_reflexive}, can be applied to binary predicates. In particular, when a predicate is marked with \texttt{symmetric}, the program recognizes \texttt{p(A,B)} and \texttt{p(B,A)} as equivalent, thus considering only one rule with this predicate. 
The constant \texttt{anti\_reflexive} prevents reflexivity, that is, predicates \texttt{p(A,A)} are not considered. 
Finally, to further refine the hypothesis space, predicates such as \texttt{\#maxv} can be used to specify the maximum number of variables occurrences in any rule generated. 
To clarify the differences between the two approaches, consider this example:
\begin{Verbatim}[breaklines=true]
% 1st approach: direct rule declarations
5 ~ damage(R, V) :- slap(R, V), agent(V), agent(R), R!=V.
% 2nd approach: mode declarations
#modeh(damage(var(agent), var(agent)), (anti_reflexive, positive)).
#modeb(1, slap(var(agent), var(agent)), (anti_reflexive, positive)).
\end{Verbatim}
In the first approach, the rule based on a prior judgment is directly encoded into the program (the number 5 is its length).
Conversely, in the last approach, the predicates that may appear in the head and body of the rule to be learned are specified without encoding the entire rule. In the example above, we do not explicitly add the constant \texttt{symmetric}, as the two variables representing agents are distinct (i.e., they refer to the roles of ``reo'', that is the offender, and ``victim''). Consequently, the hypothesis space accounts for both cases, whether the variables are symmetric or not, but the system ultimately infers from the examples provided in the program that they are distinct.

\section{The Italian Criminal Code and its encoding in ASP}\label{sect:ASPLAW}

\subsection{ASP encoding}

To create a background knowledge for ILASP, some articles of the ICC\footnote{\url{https://www.gazzettaufficiale.it/sommario/codici/codicePenale}} have been manually encoded in ASP. The articles were chosen according to our first work \cite{TALUBelgio23}, which was focused on encoding vague concepts. Vagueness is present in several articles of the ICC; for our case study, we chose to focus on the vagueness involved in the definitions of theft and theft by snatching. Another instance of vagueness, that is the level of \emph{bruise}, emerged during the encoding of articles related to assault and personal injury. ILP was not used to infer these rules, as they are immutable and must be precisely defined. For example, the rule ``\emph{if you subtract an object from another person by snatch, then you are guilty of snatch theft}'' was translated as follows:
\begin{Verbatim}[breaklines=false]
theft_snatch(R, V) :- theft(R, V, C), res_violence(R, C), 
                      not person_violence(R, V). 
\end{Verbatim}
The last literal represents the exceptions to this verdict; indeed, its meaning is that a case qualifies as snatch theft unless the snatching has caused violence toward the person, in which case the judgment changes to robbery. 

The encoding in ASP could have been made using LLMs; however, these models still exhibit weaknesses and low sensitivity to the nuances of legal language. For instance, when prompted to encode the concept of an object's physical adherence to a victim's body (see the following paragraph), ChatGPT produces an output that clearly fails to capture the precise legal meaning we intend to express:
\begin{Verbatim}[breaklines=true]    
% Tight adherence:   theft is complete as soon as 
%     the thief establishes effective control over the object.
tight_adherence(P, X) :- establishes_control(P, X).

% Tight adherence implies physical adherence but
%     focuses on the immediate establishment of control.
establishes_control(P, X) :- physical_adherence(P, X).
\end{Verbatim}
Even if the encoding was performed manually, we tried to keep homogeneity in the predicates. This means that if the same concept, such as ``violence'', appears in more than one article, then it must be encoded with the same predicate. Since legal terms have many shades, this work required also discussing with experts of the juridical field and this part was presented in \cite{TALUPolonia24}. Furthermore, also to reduce the vocabulary size, either predicates and variables (for ILASP) are inflected in the singular, as for nouns. Instead, verbs encoding criterion is to conjugate them in singular and present tense.

This part of encoding, at the beginning, comprehended only the encoding of articles, without considering any previous judgment. By the time we learned new rules from ILASP, those rules were added to the set after being ``paired'' with a marker predicate to distinguish them from the ones deriving directly from articles (see Section~\ref{sec:ilaspenc}). The reason behind this is that, in Italy, old judgments can be used just as suggestions, but they do not become part of the legal code. Adding the marker predicate allowed us to inform users whether the outcome was based on previous judgments or just on articles.

\subsection{Vagueness handling}
Vague concepts are very common in legal texts and therefore must be taken into account when encoding laws.  This study addresses vague concepts by exploiting non-deterministic choices. In other words, we assign to a vague concept a range of possible meanings to choose from. For example, as presented in \cite{TALUBelgio23}, the level of adherence of an object to the victim's body is decisive to distinguish between robbery and snatch theft. Unfortunately, this level of adhesion is not objectively describable in any way. In such cases of non-determinacy, choice rules can be applied:
\begin{Verbatim}[breaklines=true]
1{adherence(V,C,L):level(L)}1 :- unknown_adherence(V,C), 
                                 subtracted_obj(C), victim(V).
\end{Verbatim}
Those kind of rules lead to multiple answer sets, specifically one for each level of adherence. 
During the procedural trial, the tool can also be used to explain the reasoning behind a judge's decision. New evidence or information can then be integrated to refine these justifications by introducing denials. For instance, if a photograph demonstrates that there was sufficient space between the victim's handbag and their waist, the ASP constraint that follows can be added to reflect this fact:
\begin{Verbatim}
:- adherence(V,C,L), level(L), L < 2, subtracted_obj(C), victim(V).
\end{Verbatim}

\subsection{Model verification and static refining}
As result of the previous stages, we obtain an ASP program, $P_{law}$, which encodes juridical principles. 
To enhance the quality of this model, a refinement step can be performed.

A set of known criminal cases is retrieved from the databases {Foroplus} and {DeJure}.\footnote{\url{https://www.foroplus.it/home},~ \url{https://dejure.it}}
Each case $C$ is composed by a set (conjunction) of atoms. Let us focus on a case $C= \{A_1,\dots,A_k\}$, then,
\begin{itemize}
\item For $i=1,\dots,k$, run the ASP solver on the program $P_{law} \cup \{A_i\}$.
If no stable model is found (inconsistency), it means that the rules in $P_{law}$ are too restrictive. 
In such cases, the issue is examined and the program is refined accordingly by adding rules that introduce possible exceptions. This is typically achieved by relaxing strict rules through the addition of literals such as~\texttt{not exception(.)}, where \texttt{exception} is a newly introduced predicate that accounts for exceptional scenarios. 

\item Otherwise (a stable model exists), a further verification step is performed. 
It consists in adding a denial  $\leftarrow \naf A_j$ for each atom $A_j$ of $C_i$. 
The new program is input to the ASP solver to check whether the expected model is obtained. If it is not, this suggests that some rules in $P_{law}$ are too weak and again it needs refinement.
\label{SECONDOTIPO}
\end{itemize}

Once the model behaves as expected, a final verification can be conducted to determine whether different models arise when only a subset of $C_i$ facts is assumed to hold. This process can reveal potential exceptions that extend $P_{law}$, refining the legal knowledge encoded in the system.

\subsection{Contradictions' handling}
During the manual encoding process, contradictory rulings were encountered. To address these discrepancies, an initial approach was to prioritize them through the application of the so-called ``\emph{Massime di Interpretazione}'', principles designed to interpret and resolve conflicting arguments \cite{Ashley2017}. These maxims are divided into three categories:
\begin{itemize}
    \item ``\emph{Lex specialis}'': specific rulings take precedence over more general ones.
    \item ``\emph{Lex superior}'': norms issued by higher authorities are given priority.
    \item ``\emph{Lex posterior}'': recent verdicts have priority over earlier ones.
\end{itemize}
In some cases, it was not possible to establish a priority based on these principles, as they were in conflict with one another. To highlight these inconsistencies, an additional predicate, \texttt{contradiction}, was introduced. Thus, when two conflicting verdicts arise - with the facts supporting one offense while an alternative judgment suggests another - the predicates within the answer set will denote the existence of a contradiction.  
For instance, a ruling (Tribunale Bari sez. I, 26/08/2022, n. 3684) asserts that ``cervicalgia''\footnote{\textit{neck pain or cervical pain}} cannot be considered a medical condition, but merely ``pain''. This contrasts with an earlier judgment by a higher authority (Cassazione penale sez. II, 12/03/2008, n. 15420), which classifies ``cervicalgia'' to be an illness. In this case, the two rulings are in contrast because the first one should have priority due to ``Lex posterior'', while the second one should take precedence for ``Lex Superior''. In the first case, if the harm caused by the offender is identified as ``cervicalgia'', the crime cannot be categorized as ``personal injuries''. However, in the second case, it does, since ``cervicalgia'' is considered an illness. This contradiction can be addressed as follows:
\begin{Verbatim}
contradiction("not illness", "illness", I) :- only_pain(I), illness(I).
\end{Verbatim}

Another case of contradiction emerged when sentences were inconsistent with the legal articles themselves. The following example concerns the crime of personal injuries caused in conjunction with snatch theft. According to Articles 624 bis and 628, the distinction between this crime and robbery lies in the fact that, in the latter, violence or threats are directed against the victim. This violence also corresponds to the one involved in beatings and personal injuries (Tribunale Nocera Inferiore, 23/06/2020, n. 551).
Therefore, personal injuries should be related to robbery rather than snatch theft. According to the encodings of the articles, the facts activate the rule whose head is \texttt{robbery}, but the actual judgment classifies it as snatch theft. When both predicates are present, a contradiction arises, triggering the \texttt{contradiction} predicate.

\section{Predictions with explanations}\label{sect:predict}
Once the ASP solver is run on a program $P_{law}$ and a sentence $C$ added as denial (see Section~\ref{SECONDOTIPO}),
the resulting stable model can be analyzed to extract the reasoning behind the outcome. The tool can be utilized in two modes. Firstly, it can explain the reasoning behind independent judicial decisions, in order to provide a clearer understanding of how the verdicts are reached. Secondly, its capability of generating different possible consistent scenarios,  in the case of vague legal concepts, can be exploited.  

\medskip

Two approaches for ASP explainability have been investigated.
The first one makes use of \texttt{xclingo2}, which allows users to annotate rules and facts in order to get justifications. Indeed, its key strength lies in its ability to generate explanations for why certain atoms appear in an answer set, allowing rules and atoms to be described in natural language. We report an example: consider the following program
\begin{Verbatim}
cause("Carlo", "Beatrice", "skin lesion").
physical_illness("skin lesion").
intent_to_harm("Carlo", "Beatrice").
\end{Verbatim}
Examples of annotations (the ones starting with \verb|%!|) we made in \texttt{xclingo2} are the following:
\begin{Verbatim}[commandchars=\\\{\}, breaklines=true]
\textcolor{teal}{%!trace \{"% is an illness", Z\} illness(Z).}
\textcolor{teal}{%!trace \{"% is a physical illness", Z\} physical_illness(Z).}
illness(Z) :- physical_illness(Z).
\textcolor{teal}{%!trace \{"It is evident that % (perpetrator) caused injuries to % (victim)", X, Y\} injuries(X, Y).}
injuries(X, Y) :- cause_illness(X, Y), intent_to_harm(X, Y), agent(X), agent(Y).
\textcolor{teal}{%!trace_rule \{"% caused % to suffer %", X, Y, Z\}}
cause_illness(X, Y) :- cause(X, Y, Z), illness(Z), agent(X), agent(Y).
\end{Verbatim}
Those annotations produced the following explanation:
\begin{Verbatim}
|__It is evident that Carlo (perpetrator) caused injuries to Beatrice (victim)
|  |__Carlo caused Beatrice to suffer skin lesion
|  |  |__skin lesion is an illness
|  |  |  |__skin lesion is a physical illness
|  |  |__Carlo caused skin lesion to Beatrice
|  |__Carlo had general intent to harm Beatrice
\end{Verbatim}
This output represents a structured justification tree explaining why Carlo has been found guilty of causing injury to Beatrice. In fact, the tree structure represents a chain of reasoning, where each indented line explains or justifies the statement above it. The average time required to get this kind of explanations is 0.078 s (30 test cases). In particular, it emerged that explanations for a negative verdict, that is when there is no evidence of the crime, required around 0.01 s more than the positive ones. 

\begin{figure}
\centerline{\includegraphics[width=1.25\textwidth, angle=90]{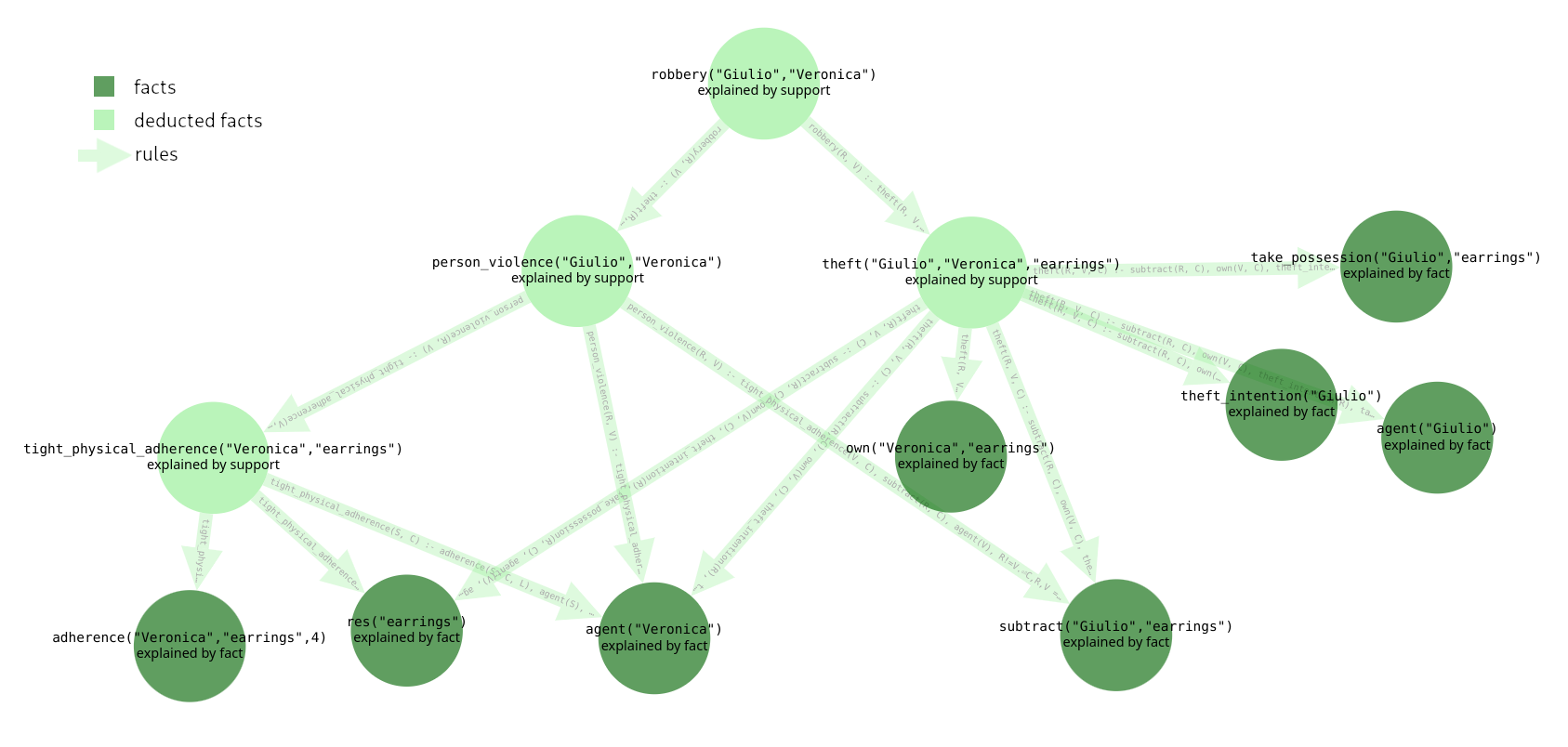}}
    \caption{An example of a DAG produced by xASP to explain inferences (cf., Section~\ref{sect:predict}). The colors are those used by xASP: dark green represents the facts, while light green indicates inferred facts.  The edges show the rule used in the inferences and indicate the facts on which they depend.}
    \label{fig:xasp}
\end{figure}

The other approach utilizes xASP, which provides a visual explanation of the outcome, representing the answer set by a DAG. 
For example, in the judgment (Cassazione penale sez. II, 21/02/2019, n. 16899), the court classified as robbery the defendant's actions of holding the victim still by pressing her head while taking her earrings. This scenario can be encoded as follows, assuming that the adherence of the object was tight since earrings are locked to the earlobes:
\begin{Verbatim}
own("Veronica", "earrings").
subtract("Giulio", "earrings").
snatch("Giulio", "earrings").
take_possession("Giulio", "earrings").
adherence("Veronica", "earrings", 4).
\end{Verbatim}

In the graph shown in Fig.~\ref{fig:xasp}, nodes represent (inferred) facts, while arrows indicate the rules applied to derive conclusions. Specifically, each arrow points to the nodes that support its inference. The average time required to get the explanation graph was of 1.65 s (30 test cases).

\section{Dynamic Model refining}\label{sect:medelRefine}
Since new sentences on crimes are made anytime, it is useful to implement a module, preferably automated, to continuously refining the tool when adding this new knowledge to $P_{law}$. We employed ILASP (Section~\ref{SUBILASP}) that learns new rules from the new sentences and extends $P_{law}$ accordingly. The manual intervention may be necessary in cases where contradictions arise.

\subsection{ILASP encoding}
\label{sec:ilaspenc}
In order to learn legal rules, there were manually encoded examples of previous judgments. The variables were instantiated according to the verdicts, taken from two legal databases {Foroplus} and {DeJure} used (see Section~\ref{sect:ASPLAW}).
The terms belong to the specific domain of the problem/crime; for example, personal injuries involve medical conditions such as ``ecchimosi'',\footnote{\textit{ecchymosis}} ``ematoma'',\footnote{\textit{hematoma}} etc. 

In each example, the inclusions specify the crimes of which the offender has been accused, and the fundamental predicates related to the verdict. The exclusions define the predicates that cannot appear in the corresponding answer set. 
Finally, the context encodes the facts regarding the specific judgment. An example of a judgment encoded in the program is reported below:
\begin{Verbatim}
#pos({beatings("R", "V"),       % inclusions
      damage("R", "V")},
     {derive_illness("V")},     % exclusions
     {harmful_intention("R").   % context
      slap("R", "V").}).
\end{Verbatim}

There were used both brave and cautious reasoning, respectively in positive (\texttt{\#pos}) and negative (\texttt{\#neg}) examples. The first ones must be included in at least one answer set, while the negative ones cannot be true in any answer set.   \\
The hypothesis space was explicitly defined encoding rules, each of them preceded by a number specifying the number of predicates involved, and the symbol $\sim$ representing the inclusion of the rule in the hypothesis space. Alternatively, there were used mode declarations, in which only the predicates that can appear in the head and in the body of the rule are specified. The rule learned from the example mentioned above is the following:
\begin{Verbatim}
damage(R, V) :- R != V; agent(R); agent(V); slap(R, V).
\end{Verbatim}

Then, the learned rules have been added to the background knowledge. To distinguish the articles of the ICC from the rules learned from previous judgments, a cardinality constraint was used. Specifically, the learned rules are encoded in such a manner that, when the body of the rule is activated, the answer set includes both the original predicate learned and a predicate \texttt{using\_judgment} which denotes that this verdict is referred to previous judgments. 

\subsection{Experimental Results}

We ran ILASP (version 4) on different sets of encoded legal articles listed below:
\begin{itemize}
    \item \textbf{Set$_{AC}$}: 583  
    \item \textbf{Set$_{CP}$}: 575, 579, 584, 588, 589, 589 bis, 59, 595, 609 bis, 610, 614 
    \item \textbf{Set$_{B, PI}$}: 581, 582 
    \item \textbf{Set$_{T, R}$}: 624, 624 bis, 628 
\end{itemize}

All tests were conducted on a system with an Intel 12th Gen Core i7-12700 processor, RAM of 16 GB and a maximum clock speed of 4.9 GHz. 
Its operating system is Ubuntu (24.04.2 LTS). The version of clingo used is 5.6.2.

For each set we encoded around 22 judgments as examples (except for the biggest set Set$_{CP}$, which involved 70 examples). The  dimension of hypothesis space, denoted by $|S|$, corresponds to the number of rules and is reported in Table \ref{table:result}. Some of these sets, marked with $^*$, employ mode declarations to generate the hypothesis space, while others rely on a predefined space. The use of mode declarations allows for a wider range of possible rules but comes at a higher computational cost. Indeed, for the sets that do not utilize mode declarations, incorporating them would render the learning process infeasible due to exceeding memory size. This challenge arises because, in the legal domain, laws are highly specific and introduce a large number of predicates, making the automatic generation of a suitable hypothesis space particularly demanding. In some cases, we adopted a hybrid approach, leveraging both techniques in a compensatory way. While some rules were manually encoded, others, typically the more general ones, were generated through mode declarations. This approach allows to balance the expressiveness of the hypothesis space with computational feasibility.

\begin{table}[h]
\caption{Timing breakdown (in seconds) for different ILASP stages across datasets}
\centerline{
\begin{tabular}{lrrrrr}
\hline
& \textbf{Set$^*_{AC}$} & \textbf{Set$_{CP}$} & \textbf{Set$_{B, PI}$} & \textbf{Set$^*_{B, PI}$}  & \textbf{Set$_{T, R}$} \\
\hline
$|S|$ & 5418 & 531 & 21 & 93996 & 11 \\[1.2ex]
Pre-processing & 0.016 & 0.028 & 0.023 & 0.046 &  0.014 \\
Hypothesis space generation & 0.294 & 0.102 & 0.009 & 204.179 & 0.005 \\
Conflict analysis & 4.289 & 1.876 & 0.087 & 291.992 & 0.076 \\
Counterexample search & 0.048 & 0.676 & 0.099 & 0.179 & 0.088 \\
Hypothesis search & 1.372 & 0.311 & 0.022 & 52.092 & 0.010 \\
\hline
\textbf{Total} & \textbf{6.075} & \textbf{3.005} & \textbf{0.240} & \textbf{549.576} & \textbf{0.196} \\
\hline
\label{table:result}
\end{tabular}}
\end{table}

The results shown in Table \ref{table:result} highlight the significant computational cost of mode declarations. For instance, in Set$^*_{B, PI}$, hypothesis space generation and conflict analysis require substantially more time (204.18s and 291.99s, resp.) compared to predefined rule-based configurations. In contrast, predefined rule-based sets, such as Set$_{B, PI}$ and Set$_{T, R}$, exhibit much lower execution times across all stages, demonstrating better efficiency but at the cost of reduced rule flexibility.

\section{Conclusions}\label{Conclusions}

\begin{figure}
    \centering
        \begin{subfigure}[b]{0.48\textwidth}
            \centering
            \includegraphics[width=\linewidth]{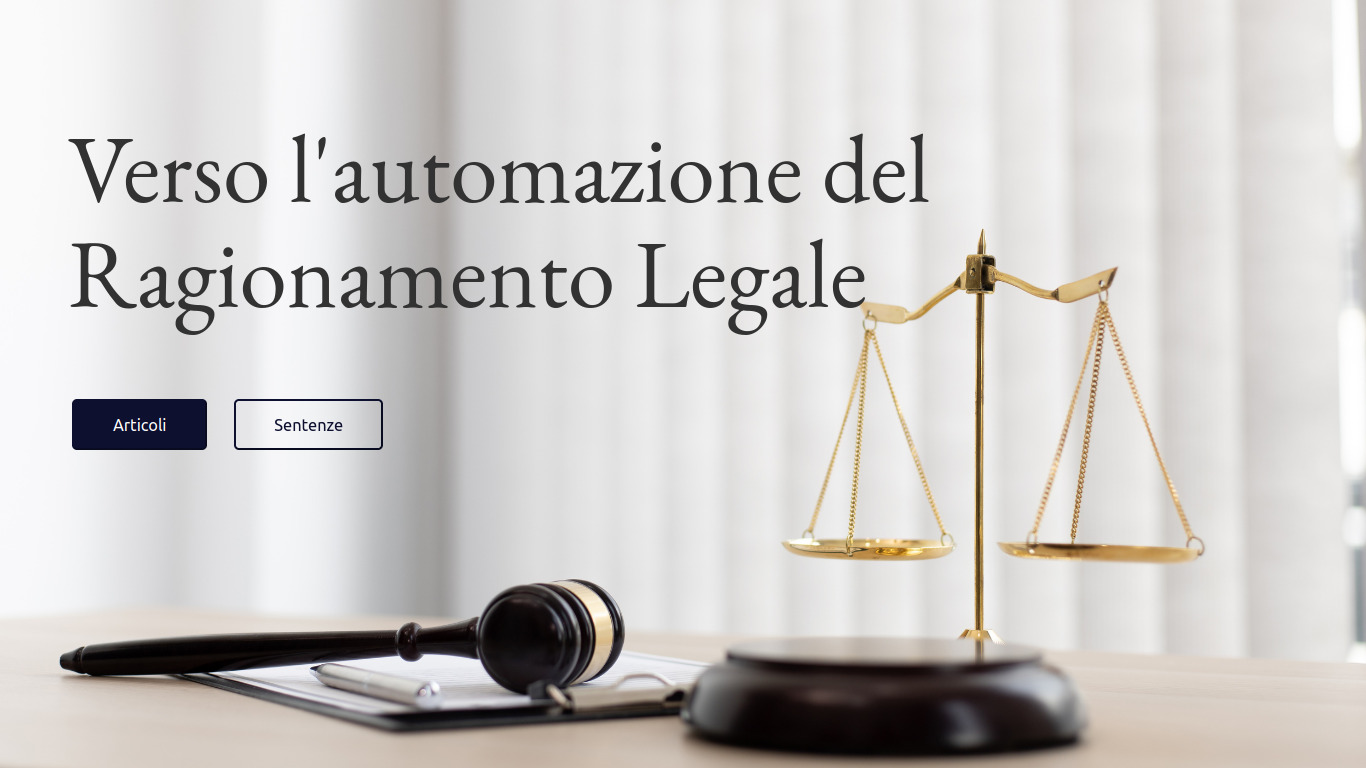}
        \end{subfigure}
        \hfill
        \begin{subfigure}[b]{0.48\textwidth}
            \centering
            \includegraphics[width=\linewidth]{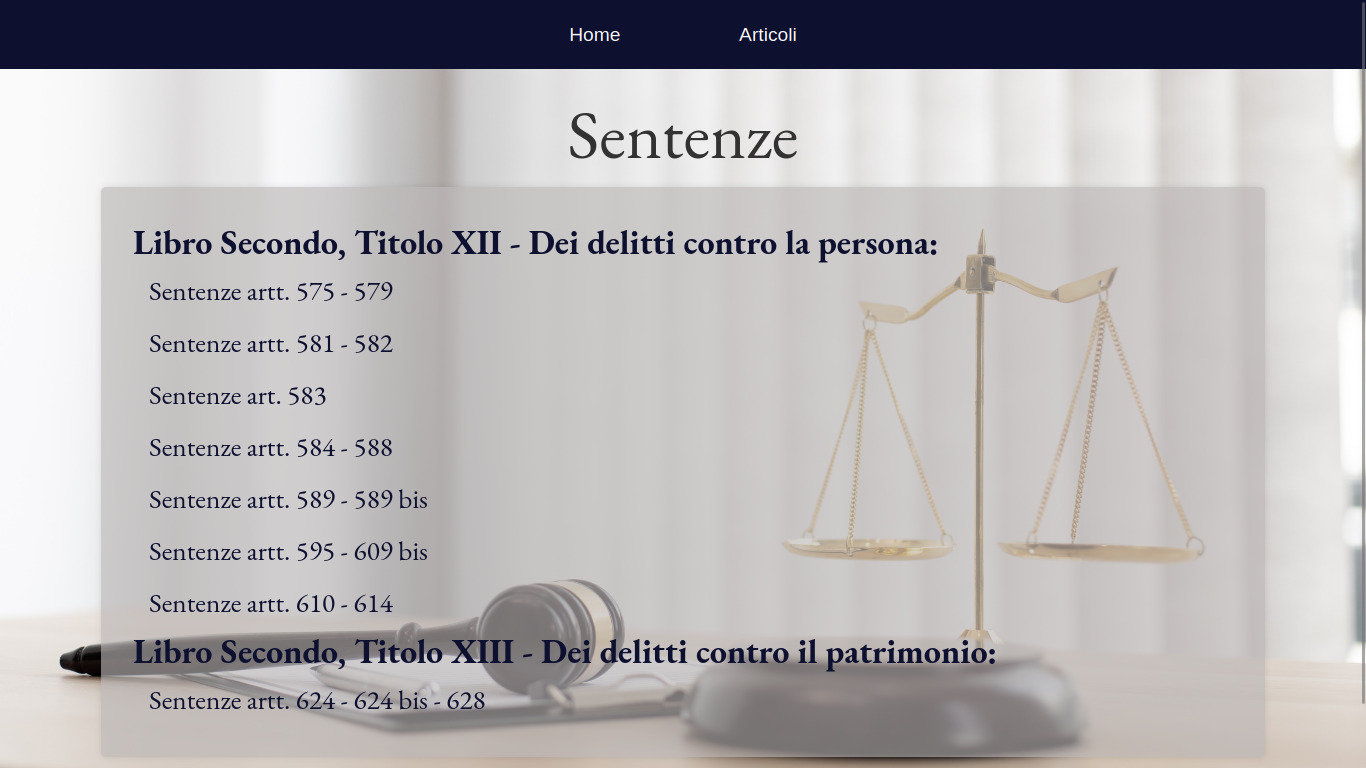}
        \end{subfigure}
        \caption{Homepage of the website (left), judgments of the encoded articles (right)}
        \label{Fig.1j}
\end{figure}

In this work, we have analyzed how to encode a legal knowledge base in ASP, how to use it for legal reasoning and for explaining the decisions made, and how to refine it almost automatically. The developed system is free and available from \url{http://clp.dimi.uniud.it/sw/}. Two screenshots are reported in Figure \ref{Fig.1j}.
The main page provides links to articles and judgments (in Italian), with each article's ASP encoding presented alongside its textual counterpart.
Examples of previous verdicts analyzed by ILASP and then incorporated in the knowledge base are organized according to the articles they pertain to. 
\medskip

The decisions made by courts of the Italian criminal system are made at three distinct levels: 1. \emph{primo grado}, 2. \emph{appello}, 3. \emph{cassazione}. At each level, the ruling can be modified depending on new proofs or by checking procedural errors of the previous level. 
We have developed a simple tool that rewrites the ASP encoding from the first level when new evidence is presented. This tool simulates the second level by updating facts. As future work we will deal with the third level.

The main advantage in adopting logic programming, compared to, for example, using ML-based techniques, is the immediate explainability properties of the system.
Moreover, in the legal domain the size of the data can still be handled by an ASP program (this is different from other domain families such as, for example, the weather forecast domain, where data is computed, stored, and analyzed for each second and for all cubic meters of the atmosphere of the planet earth). 

A judge might be interested in analyzing all stable models obtained. If a property holds in all stable models, it is reasonable to believe in it, thus with a sort of cautious reasoning, it can be proposed and accepted by humans. If it holds in one/few stable models, it might be temporarily accepted (a sort of brave reasoning), but maybe it requires further analysis. 

The choice of ILASP as inference engine for automatizing this part opens further lines of research. For instance, we intend to apply to the ILASP solver the ASP-technology developed in parallelizing ASP-solvers~\cite{DBLP:journals/tplp/DovierFGHPR22,DovierFP18,DovierFPV16,DovierFV19}
in order to improve efficiency and scalability. 
Concerning the explainability issues we will relate our results to those produced by argumentation-based approaches \cite{DBLP:journals/ai/PrakkenS15}.

On the other hand, a key limitation is that, although ILASP is capable of handling dynamic updates, the initial knowledge representation still relies on human expertise to translate legal texts into ASP rules. We acknowledge that this manual encoding process can become a bottleneck, and we therefore plan to address this in future work by exploring methods for its automation.
\bibliographystyle{eptcsini}
\bibliography{AILAW}
\end{document}